# ArSciKids: Arabic Curriculum-Aligned Datasets for Scientific Experiment and Multiple-Choice Question Generation in Primary Education

Taghrid Housari[+], Massa Rankoussi[+], Jafar Abdul Majeed[+], Wael Tarabishi,
Khloud Al Jallad*, Nada Ghneim

*Arab International University, Daraa, Syria*

**Corresponding Author*: k-aljallad@aiu.edu.sy

## Abstract

Despite growing interest in Arabic natural language processing, curriculum-aligned educational resources remain scarce compared with English, and few public datasets for Arabic curriculum content with structured, generation-ready annotations, particularly for primary-school science. We introduce ArSciKids, a collection of two Arabic datasets designed for curriculum-aligned educational content generation: (1) a scientific-experiments dataset containing 999 manually curated records from official primary-school science textbooks in Saudi Arabia, Syria, and Jordan, and (2) an assessment-question dataset containing 2,167 multiple-choice questions collected from the Saudi Manhaji educational platform. The experiment records are structured with titles, objectives, tools, procedures, observations, conclusions, keywords, and grade levels, while the assessment records contain lesson references, questions, four answer options, correct answers, and explanations. Both datasets cover grades 1–6 and consist of real-life Arabic educational content rather than machine-translated or synthetically generated material. We assess the usefulness of these resources for generation by fine-tuning ten open models spanning sub-billion to several-billion parameter scales using LoRA. Models are evaluated using automatic metrics, including ROUGE and BLEU, answer accuracy for multiple-choice generation. Moreover, we evaluated results by five-criterion human evaluation protocol. The results show substantial differences across model scales and tasks. In particular, Gemma-3-4B achieves ROUGE-1 of 0.9664 and BLEU of 0.4449 on the grades 1–2 experiment-generation split, while ROUGE-L reaches 0.4647 for multiple-choice question generation, with answer accuracy reaching 100% under the evaluated setting. Models below 2B parameters consistently perform worse across the generation tasks, suggesting that model capacity is an important factor for effective fine-tuning on these curriculum-aligned Arabic resources. ArSciKids will be publicly available and we hope that it provides a manually curated, structured foundation for research on Arabic educational NLP, curriculum-aware generation, and low-resource language modeling.

**Keywords:** Arabic NLP; educational NLP; low-resource languages; curriculum-aligned datasets; primary education; science education; scientific experiment generation; multiple-choice question generation.

## I. INTRODUCTION

Large language models (LLMs) have the potential to support teachers in preparing lessons, generating assessments, and adapting educational material to different student levels. However, the quality of such applications depends substantially on the availability of high-quality, domain-specific, and curriculum-aligned training data. For Arabic, despite the large number of speakers and learners, publicly available educational datasets remain limited, particularly for primary-school science. This limitation is especially important at the primary level, where educational terminology, linguistic complexity, instructional structure, and curriculum content can vary considerably across countries. Existing Arabic resources also provide limited coverage of structured, generation-oriented educational content derived directly from national curricula.

In this work, we focus on the data resources required for curriculum-aligned Arabic educational generation. We introduce ArSciKids[1] that contains two Arabic datasets covering primary-school grades 1–6: a scientific-experiments dataset, ArSciKids_Exp, designed to support experiment generation, and an assessment-questions dataset, ArSciKids_QA, designed to support multiple-choice question (MCQ) generation. The datasets were manually compiled from official national-curriculum textbooks and educational platforms in Arabic rather than being translated from English-language resources. Each dataset is represented in a structured JSON format with explicit fields designed for downstream supervised fine-tuning and generation. The experiment dataset captures information such as objectives, tools, procedures, observations, and conclusions, while the assessment dataset includes lesson references, questions, answer options, correct answers, and explanations.

The contributions of this paper are: introducing two curriculum-aligned Arabic educational datasets for primary-school science, and conducting baseline experiments using ten open models with LoRA-based supervised fine-tuning to evaluate the utility of the datasets for Arabic educational generation and to examine the effect of model scale on performance. Importantly, the datasets constitute the central contribution of the work; the models are used primarily as validation instruments for assessing whether the proposed resources provide sufficient signal for downstream generation tasks.

The rest of this paper is organized as follows. Section II reviews related works. Sections III describes the dataset construction. Section IV presents the datasets statistics. Experiments are described in Section V. Section VI shows results and discussion. Section VII presents human evaluation. Section VIII presents the conclusion.

## II. RELATED WORKS

### *A. Arabic STEM and Science Resources*

Native Arabic STEM resources are limited. 3LM [1] introduced three benchmarks for Arabic STEM and code, including native multiple-choice questions collected from real Arabic worksheets and textbooks across several Arab countries, plus synthetic and translated subsets. Beyond 3LM, two recent Arabic benchmarks address science knowledge more directly. AraSTEM [2] is a native Arabic benchmark of 11,637 multiple-choice questions spanning mathematics, physics, chemistry, biology, computer science, and medicine at several educational levels; its results show that publicly available models struggle with scientific Arabic and highlight the need for localized resources. ArabicMMLU [3] is the first multi-task language-understanding benchmark for Arabic, comprising 14,575 Modern Standard Arabic multiple-choice questions over 40 tasks, collected from school exams in North Africa, the Levant, and the Gulf, including natural science and physics subjects; even the best Arabic-centric model it evaluates reaches only 62.3%. Both resources are designed mainly to evaluate what models know. Neither provides the structured experiment procedures (tools, steps, observations, conclusions) or the lesson-grounded questions with explanations that are needed to train and assess curriculum-aligned generation, which is the gap ArSciKids targets.

[1] https://huggingface.co/datasets/MassaRankoussi/ArSciKids/

### B. Educational Content and Question Datasets

Neural question generation has historically been trained on English datasets such as SQuAD [4] and FairytaleQA [5] , and recent systems study Bloom-aligned question quality [6] . These resources are English and general-domain. To our knowledge, no public dataset pairs Arabic primary-school lesson content with grade-labelled MCQs, correct answers, and explanations in a single generation-ready schema, which motivates the assessment-questions dataset described here.

### C. Scientific and Physical Reasoning Benchmarks

English science benchmarks are considerably more mature. The AI2 Reasoning Challenge (ARC) [7] collects 7,787 grade-school science exam questions and separates a Challenge set that retrieval and word co-occurrence baselines fail to answer. ScienceQA [8] extends this line to roughly 21,000 multimodal K-12 science questions annotated with lectures and explanations, showing that explicit reasoning chains can improve science question answering. At a more advanced level, SciBench [9] targets college-level problems in mathematics, chemistry, and physics; the best overall score it reports is only 43.22%, and its error analysis categorizes failures into ten problem-solving skills. Together, these works show that scientific content in physics and chemistry remains a demanding test for language models even in English, and that fine-grained, human-centered error analysis is as informative as aggregate accuracy, a principle we also follow in our human evaluation (Section VII).

For Arabic, the closest work on physical reasoning is HADS [10] , a parallel Arabic benchmark of 15,977 items translated from PIQA and culturally adapted by five bilingual translators, which evaluates how everyday objects behave and how physical tasks should be performed. On its common test split, the best fine-tuned transformer (mDeBERTa-v3-base) reaches 74.35% accuracy while GPT-4o reaches 93.50% with three-shot prompting, and the error analysis identifies surface bias and lexical ambiguity as the dominant failure modes. HADS and ArSciKids are complementary: HADS measures everyday physical commonsense through binary-choice items, whereas ArSciKids provides curriculum-grounded scientific content, namely experiment procedures and multiple-choice questions with explanations drawn from primary-school textbooks and a national educational platform. The growing number of Arabic resources in physics, chemistry, and science education reflects increasing interest in this area, yet, to our knowledge, none offers grade-labelled, generation-ready records derived from primary-school curricula, which motivates the datasets presented in this paper.

## III. DATASETS CONSTRUCTION

This section describes the construction of our created datasets, ArSciKids, a collection of two Arabic educational datasets introduced in this work: the Scientific Experiments Dataset, ArSciKids_Exp, and the Assessment Questions Dataset, ArSciKids_QA. Although the datasets target different generation tasks, both were constructed with the same main objective: to provide structured, curriculum-aligned Arabic resources derived from authentic primary-school educational material. The construction process therefore focused on source authenticity, structured representation, manual verification, and consistency across records.

### A. Data Sources

The Scientific Experiments Dataset was constructed from official primary-school science textbooks covering grades 1–6 in Saudi Arabia, Syria, and Jordan. The source material was obtained from three educational repositories: the Mnhaji platform [11] for the Saudi curriculum, the Syrian Ministry of Education Curricula Portal [12] for the Syrian curriculum, and the Al-Mazeyd educational platform [13] for the Jordanian curriculum. In total, 18 PDF textbooks were studied.

The Assessment Questions Dataset was constructed primarily from the Saudi curriculum through the Mnhaji platform [11] . The main source consisted of lesson summaries containing lesson content and associated assessment questions. Because lesson summaries were less consistently available for the earliest grades, supplementary material was collected from grades 1–2 textbook chapters 1–2. The two sources were subsequently integrated and de-duplicated to reduce overlapping records.

The use of multiple national curricula for the experiment dataset and a consistent national curriculum for the assessment dataset serves complementary purposes. The former introduces variation in scientific terminology, instructional phrasing, and curriculum organization, whereas the latter provides a relatively homogeneous assessment style and answer format.

### B. Collection and Annotation Procedure

Data collection was performed through a manual, schema-driven process rather than by translating or synthetically generating English-language educational content. Source documents were first inspected to identify relevant scientific experiments, lesson content, and assessment questions. The identified material was then transferred into predefined JSON schemas. The collection process was designed to preserve the instructional content of the original sources while converting heterogeneous textbook and platform representations into a common machine-readable format. Records were manually checked during extraction to ensure the correctness of it. The datasets collection required approximately two months of work.

As for the JSON schemas (see Figures 1, 2), the experiment record contains eight fields: title, objective, tools, steps, observations, conclusion, keywords, and grade. The tools, steps, and keywords fields are represented as lists, while the remaining fields contain textual values. The assessment-question record contains eight corresponding fields: lesson_title, lesson_content, question, options, correct_answer, explanation, and grade, with the four answer choices represented as a list within the options field. The explicit separation of questions, answer options, correct answers, and explanations makes the dataset directly applicable to conditional generation and evaluation of MCQ generation systems.

Samples from both datasets are shown in Figures 3,4.

```
{
  "title":        "...",
  "objective":    "...",
  "tools":        ["...", "..."],
  "steps":        ["...", "..."],
  "observations": "...",
  "conclusion":   "...",
  "keywords":     ["...", "..."],
  "grade":        "2"
}
```

*Figure 1: Schema of a Scientific Experiment Sample*

```
{
  "lesson_title":   "...",
  "lesson_content": "...",
  "question":       "...",
  "options":        ["د","ج","ب","أ"],
  "correct_answer": "...",
  "explanation":    "...",
  "grade":          "2"
}
```

*Figure 2: Schema of an Assessment Question Sample.*

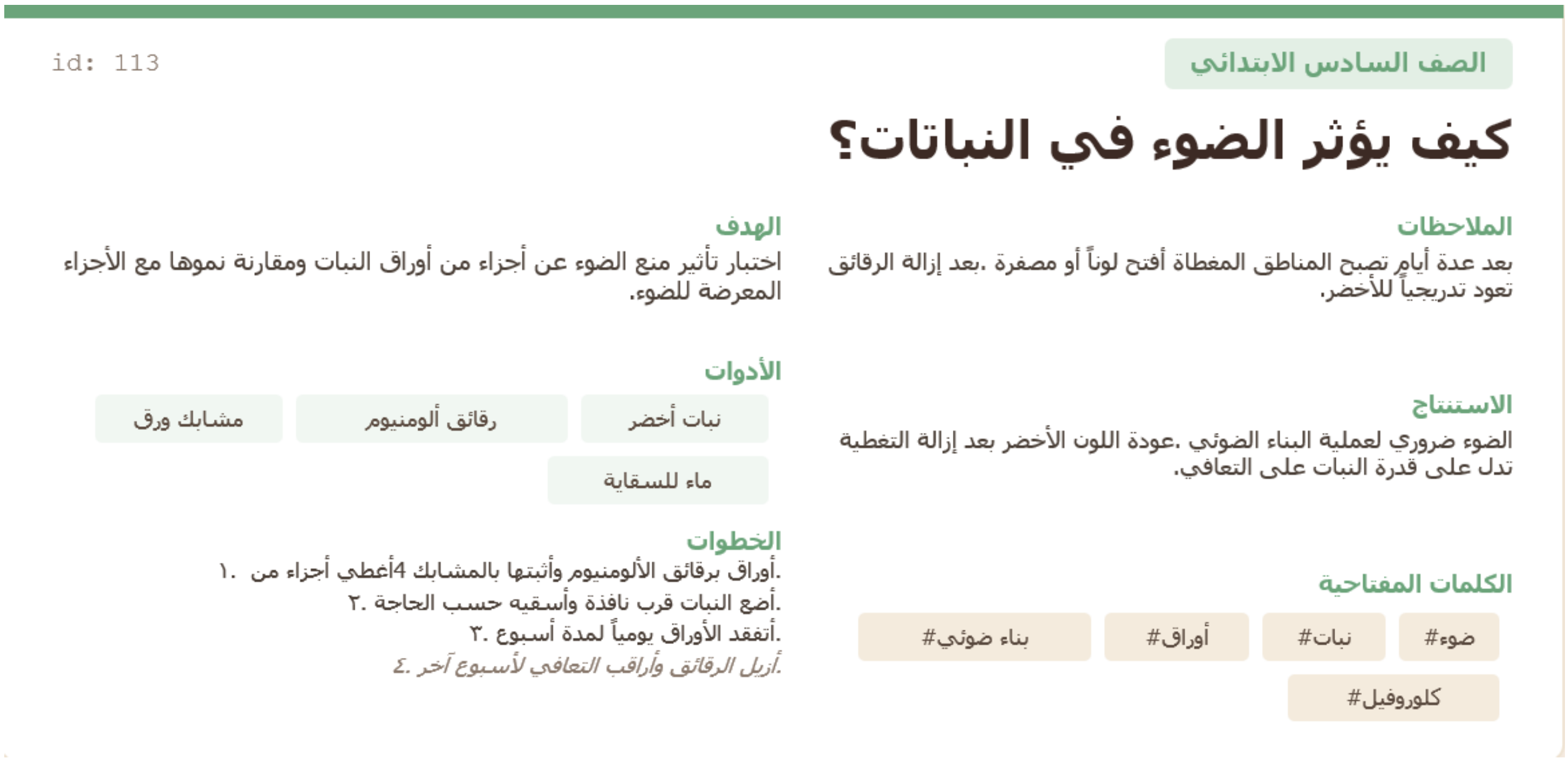

*Figure 3: Scientific Experiments Dataset Sample*

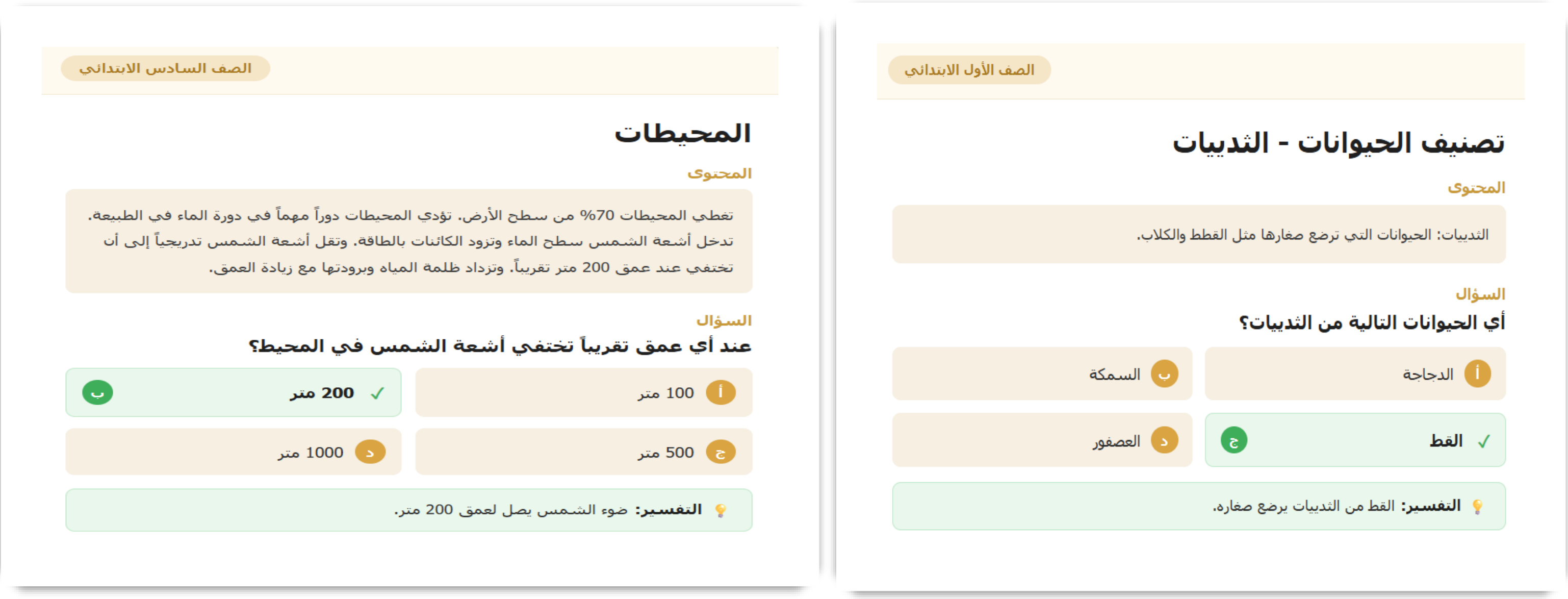

*Figure 4: MCQ Dataset Samples*

## IV. DATASET STATISTICS

The final resources contain 999 scientific-experiment records and 2,167 assessment-question records, covering grades 1–6. Their overall characteristics and linguistic distributions are analyzed in this section. Figure 5 shows the number of samples in each grade for each of the two created datasets, where strongest concentration on grades 4 and 6.

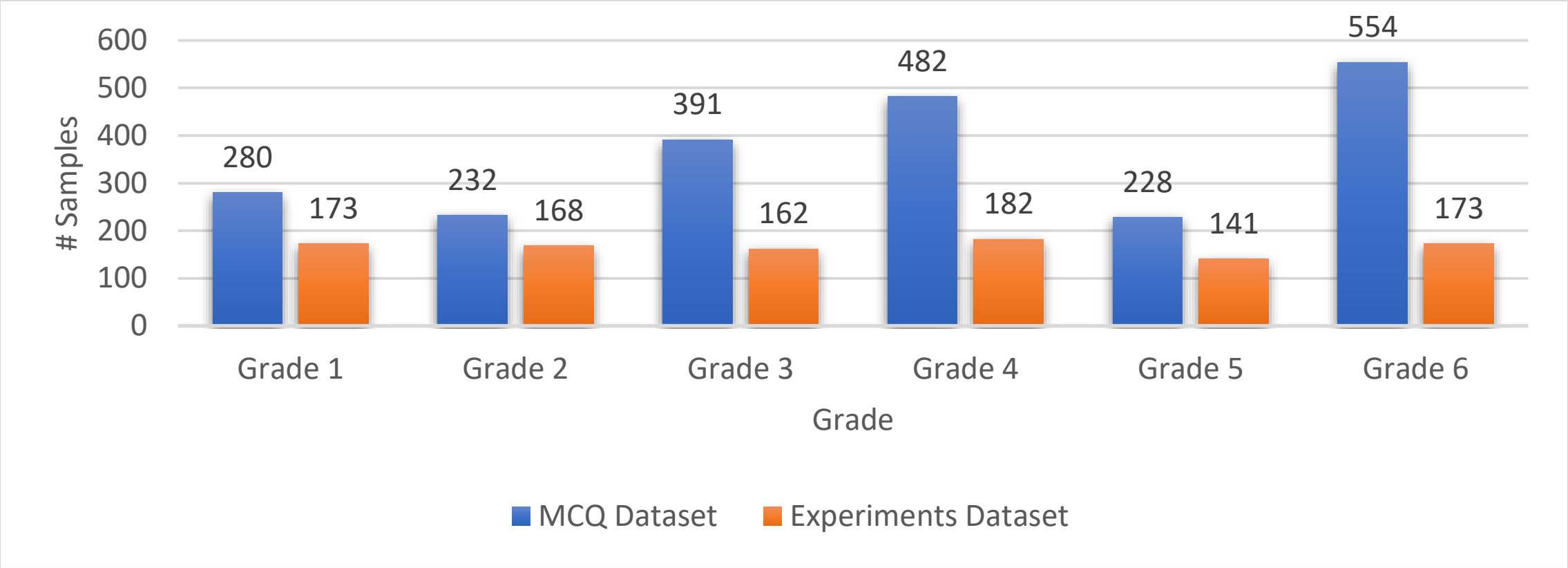

*Figure 5: ArSciKids Datasets Samples Counts*

Figure 6 presents the domain distribution of the Scientific Experiments and Multiple-Choice Questions (MCQs) datasets, where Biology is the dominant domain in both datasets, while the distribution of the remaining domains varies.  As for Experiments dataset: Biology accounts for the largest share (26%), followed by Physics (25%) and Environment (24%). Matter represents 18%, while Other (5%) and Health (2%) constitute smaller portions. As for Questions dataset: Biology is again the largest domain (36%), followed by Environment (24%) and Matter (12%). Other accounts for 16%, Physics for 9%, and Health for 3%.

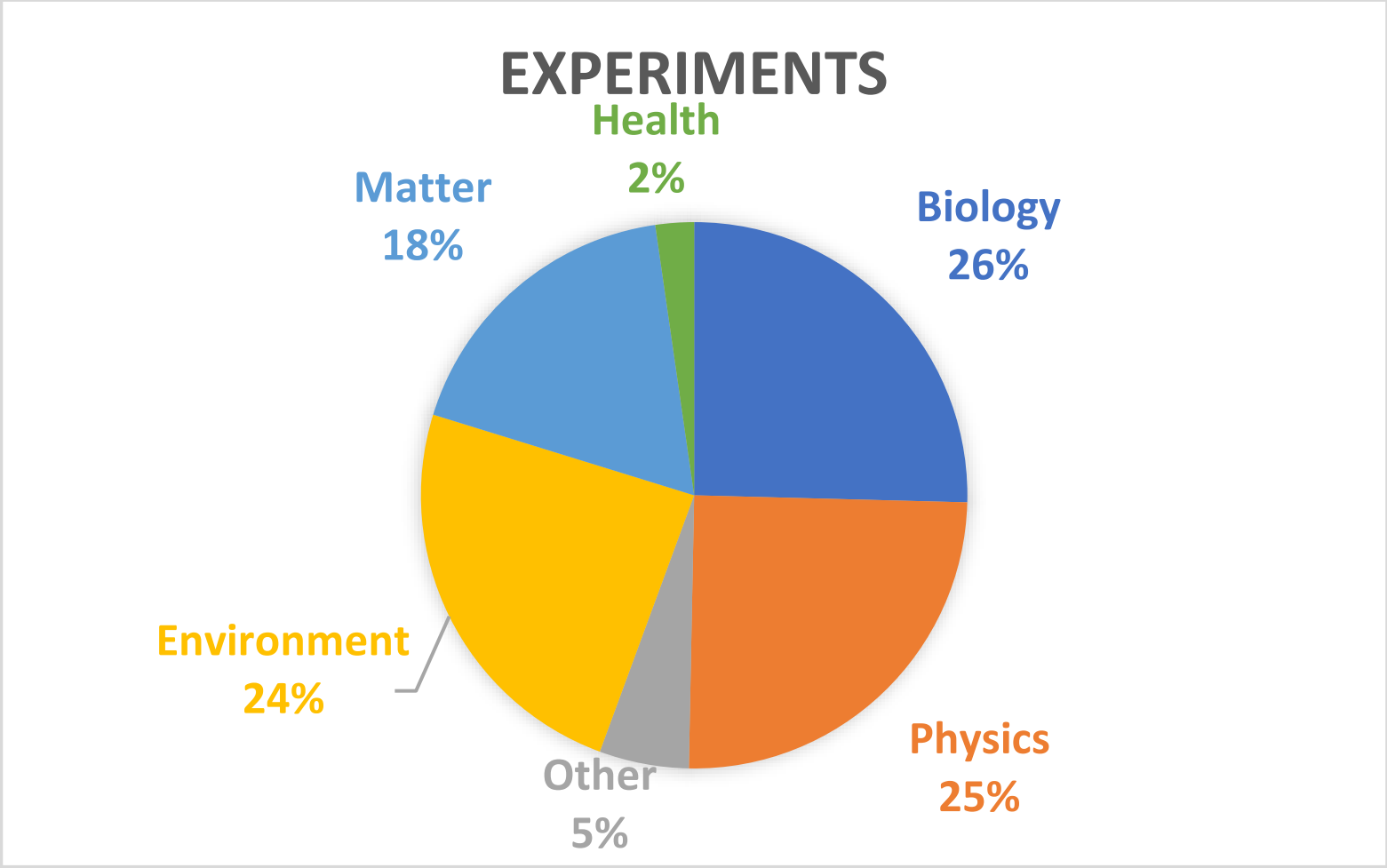

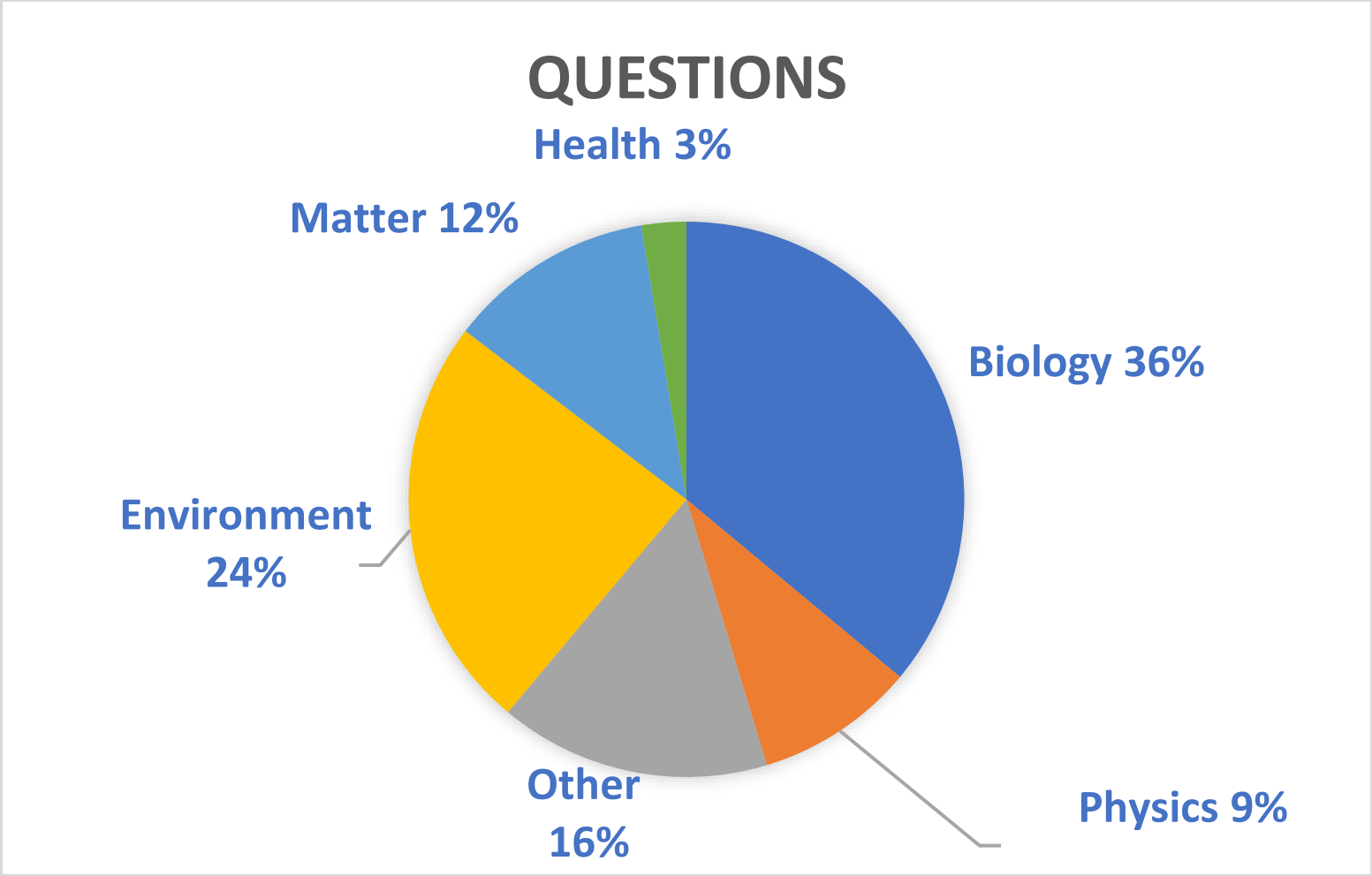

*Figure 6: Scientific Experiment & MCQs Datasets Domain Distribution*

Tables 1 and 2 summarize the word-level statistics of the Experiments and MCQ datasets, respectively. The Experiments dataset contains 999 instances, with an average of 62.47 words per instance and 379.43 characters. Each experiment consists of an average of 5.43 procedural steps and 4.84 tools, with approximately 6.00 keywords and 5.20 title words. The objective, observation, and conclusion sections contain an average of 9.20, 20.40, and 27.66 words, respectively, indicating that conclusions and observations constitute substantial portions of the experimental descriptions. In comparison, the MCQ dataset contains 2,167 instances, with an average of 36.05 words per instance and 223.29 characters. Each question contains approximately 1.98 content sentences, 6.08 question words, and 2.94 answer words, while explanations and answer options contain an average of 6.71 and 8.73 words, respectively. Overall, the statistics indicate that the Experiments dataset contains substantially longer and more structurally detailed textual descriptions, whereas the MCQ dataset is more concise and centered around question–answer structures.

| | **Count** | **Mean** | **Std** | **Min** | **25%** | **50%** | **75%** | **Max** |
|---|---|---|---|---|---|---|---|---|
| **Totals Words** | 999.0 | 62.47 | 30.37 | 13.0 | 41.0 | 55.0 | 77.0 | 195.0 |
| **Total Chars** | 999.0 | 379.43 | 189.72 | 72.0 | 247.5 | 335.0 | 468.5 | 1244.0 |
| **Procedure Steps** | 999.0 | 5.43 | 2.01 | 2.0 | 4.0 | 5.0 | 6.0 | 19.0 |
| **Tools Count** | 999.0 | 4.84 | 2.23 | 1.0 | 3.0 | 5.0 | 6.0 | 19.0 |
| **Keywords Count** | 999.0 | 6.00 | 1.95 | 1.0 | 5.0 | 6.0 | 7.0 | 14.0 |
| **Title Words** | 999.0 | 5.20 | 2.06 | 1.0 | 4.0 | 5.0 | 6.0 | 13.0 |
| **Objective Words** | 999.0 | 9.20 | 3.20 | 3.0 | 7.0 | 9.0 | 11.0 | 24.0 |
| **Observations Words** | 999.0 | 20.40 | 12.00 | 2.0 | 12.0 | 18.0 | 25.0 | 78.0 |
| **Conclusion Words** | 999.0 | 27.66 | 17.39 | 3.0 | 14.0 | 23.0 | 37.0 | 101.0 |

*Table 1: Experiments Dataset Words Statistics*

| | **Count** | **Mean** | **Std** | **Min** | **25%** | **50%** | **75%** | **Max** |
|---|---|---|---|---|---|---|---|---|
| **Total Words** | 2167.0 | 36.05 | 10.65 | 15.0 | 29.0 | 33.0 | 42.0 | 99.0 |
| **Total Chars** | 2167.0 | 223.29 | 65.45 | 88.0 | 176.0 | 209.0 | 257.0 | 574.0 |
| **Lesson Title Words** | 2167.0 | 2.54 | 1.10 | 1.0 | 2.0 | 2.0 | 3.0 | 7.0 |
| **Content Words** | 2167.0 | 17.78 | 9.41 | 3.0 | 11.0 | 15.0 | 22.0 | 71.0 |
| **Content Sentences** | 2167.0 | 1.98 | 1.42 | 1.0 | 1.0 | 2.0 | 2.0 | 12.0 |
| **Questions Words** | 2167.0 | 6.08 | 2.30 | 2.0 | 4.0 | 6.0 | 7.0 | 21.0 |
| **Answer Words** | 2167.0 | 2.94 | 1.87 | 0.0 | 1.0 | 2.0 | 4.0 | 11.0 |
| **Explanation Words** | 2167.0 | 6.71 | 2.75 | 3.0 | 5.0 | 6.0 | 8.0 | 26.0 |
| **Options Words** | 2167.0 | 8.73 | 4.04 | 0.0 | 5.0 | 8.0 | 11.0 | 26.0 |

*Table 2: MCQ Dataset Words Statistics*

Detailed lexical statistics are shown in Figures 7, 8 for ArSciKids_QA, ArSciKids_Exp, respectively. Moreover, Figures 9, 10 shows the most frequent words in each dataset.

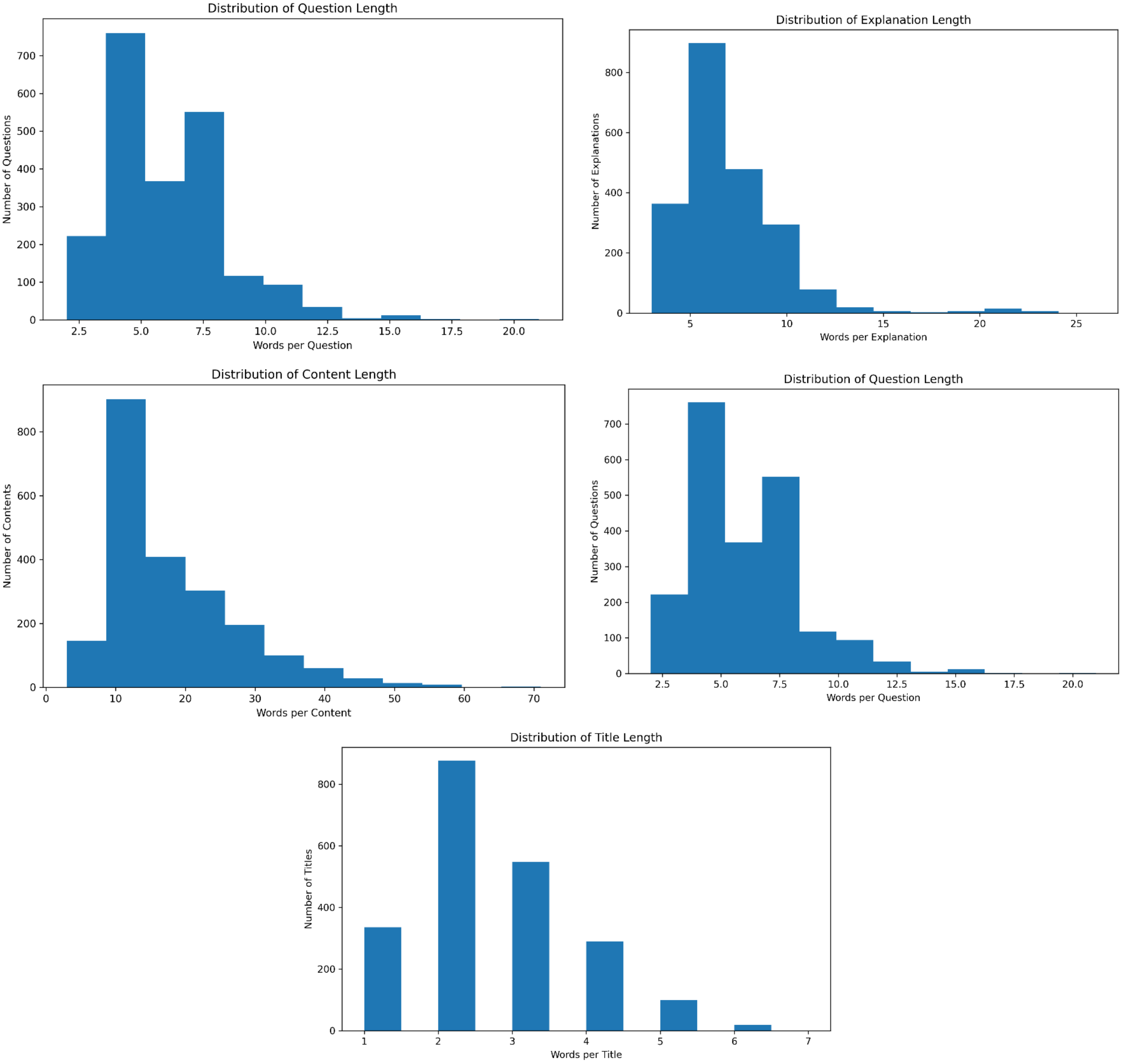

*Figure 7: Detailed Lexical Statistics of ArSciKids_QA*

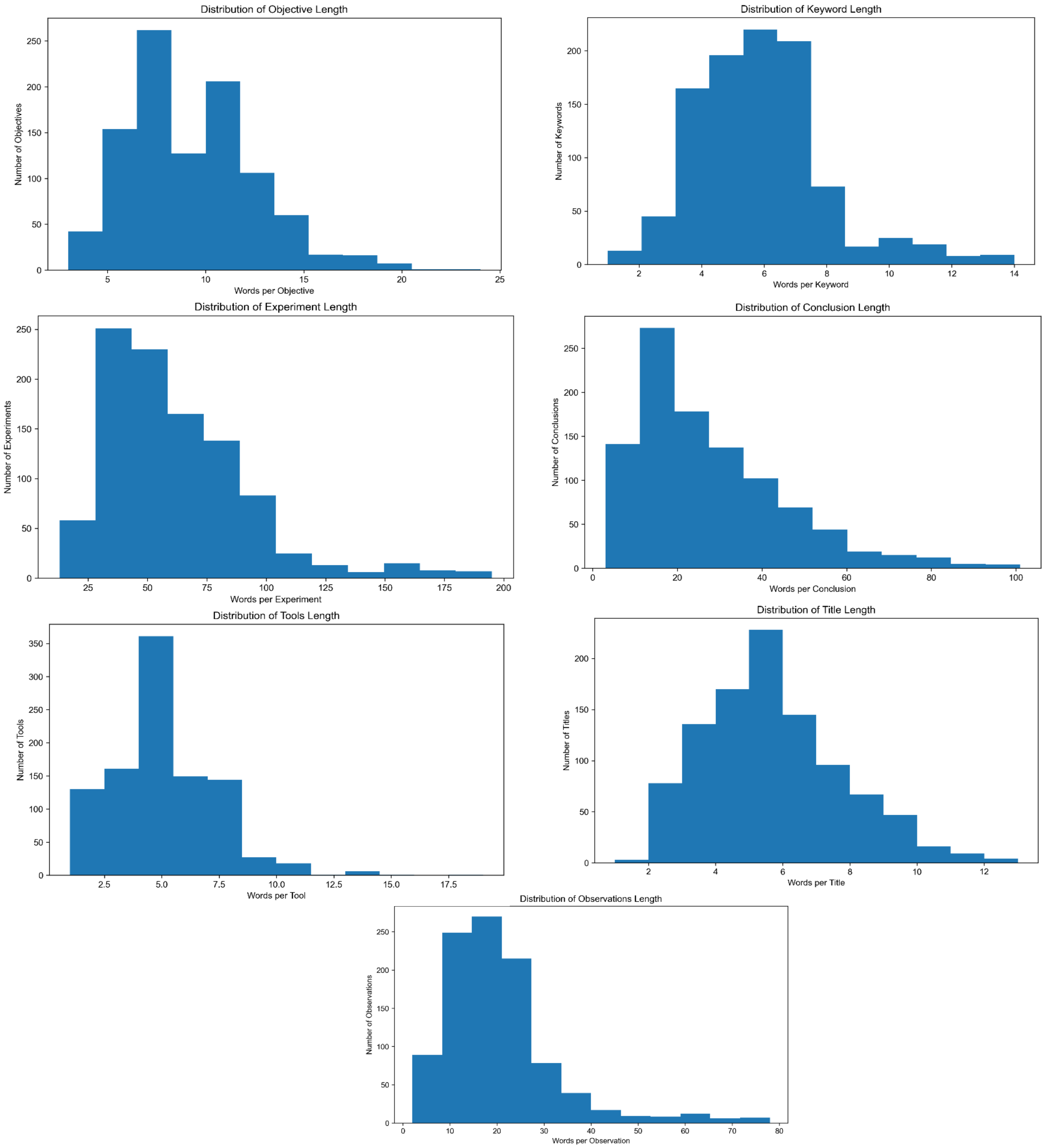

*Figure 8: Detailed Lexical Statistics of ArSciKids_Exp*

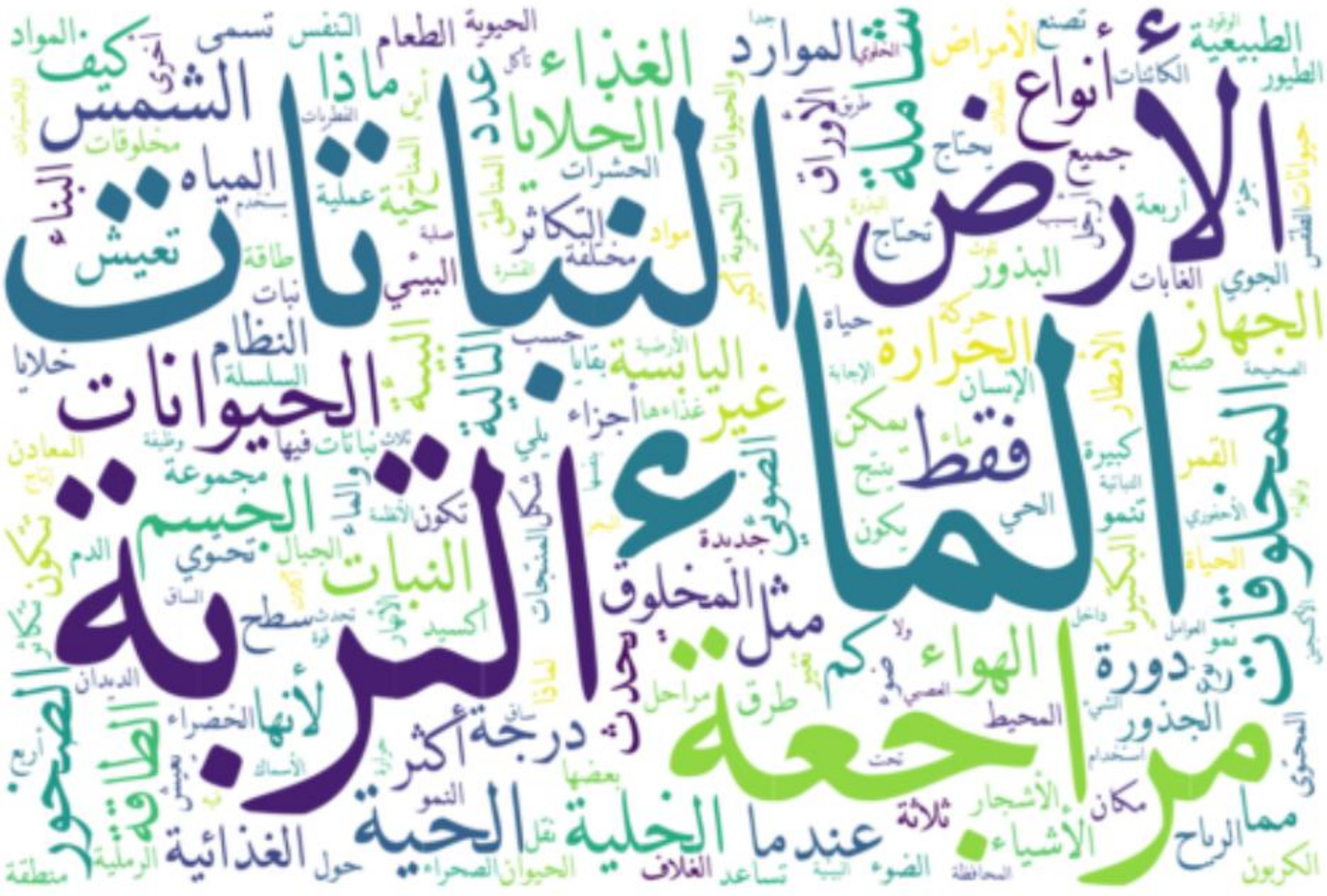

*Figure 9: ArSciKids_QA Dataset Word Cloud*

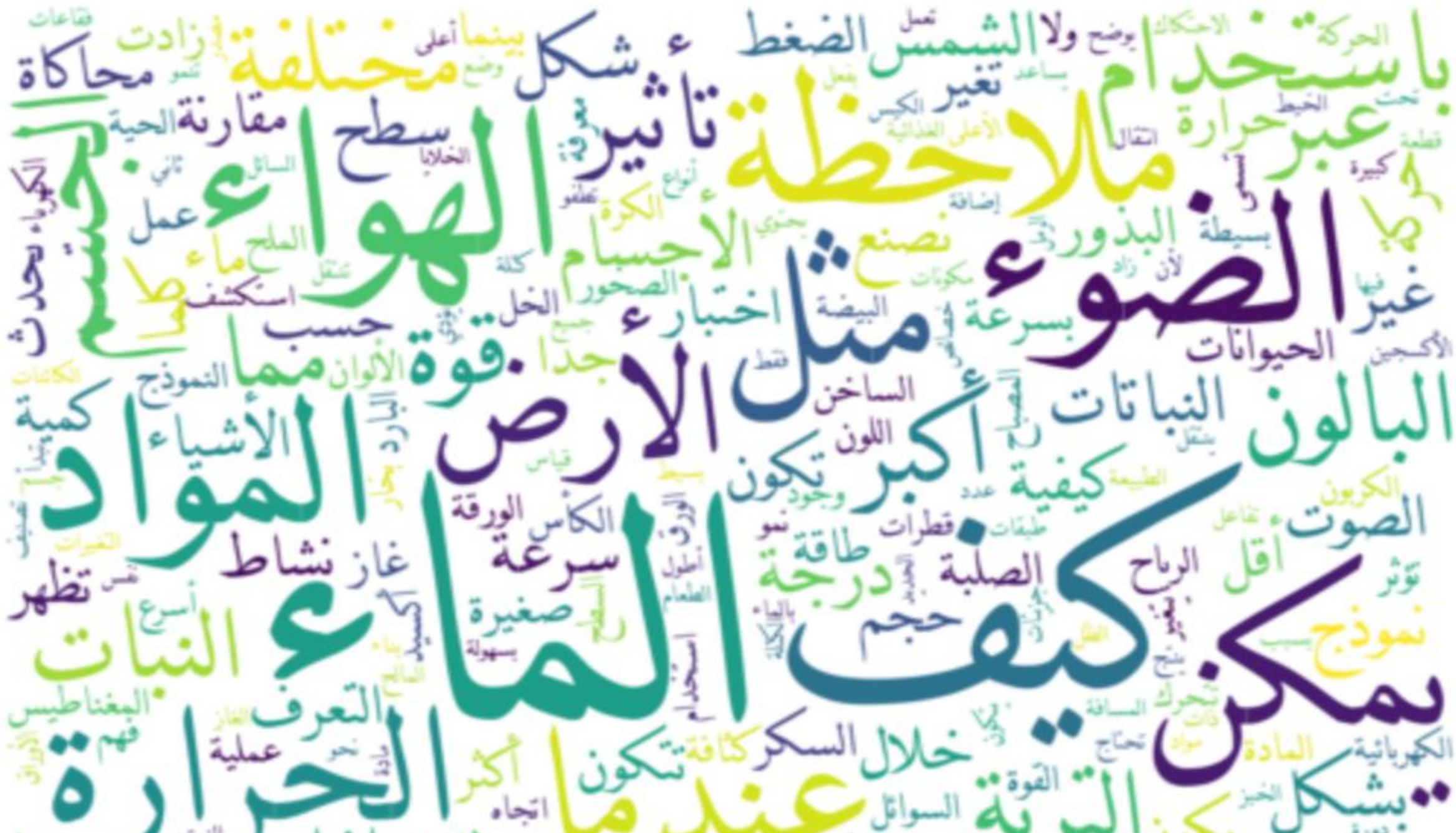

*Figure 10: ArSciKids_Exp Dataset Word Cloud*

## V. EXPERIMENTS

To evaluate our created datasets, ten models were fine-tuned with Low-Rank Adaptation (LoRA) and evaluated.

### A. PREPROCESSING

Several configurations were compared during fine-tuning:

- Without any preprocessing (raw text)
- With basic Normalization

### B. USED MODELS

As for Experiments dataset, we used the following models:

- Gemma-3-4b[14]
- MedGemma-4b[15]
- Falcon3-1B [16]
- Qwen1.5-1.8B[17]
- Falcon-H1-0.5B[18]
- BLOOM-560M/1.7B[19]
- StableLM-2-1.6B[20]
- TinyLlama-1.1B[21]

As for MCQ dataset, we used the following models:

- Gemma-3-4b[14]
- MedGemma-4b[15]
- Gemma-2-2B[22]
- Qwen1.5-1.8B[17]

### C. EXPERIMENTS SETUP

LoRA[23] was applied to the (q,k,v,o,gate,up,down) modules, typically r=16/α=32, dropout 0.05, max length 1024 tokens, learning rate 1e-4 to 2e-4, with chat templating, completion-only loss masking, dynamic padding, and early stopping. Automatic metrics were ROUGE [24] and BLEU [25]. answer selection used letter accuracy.

## VI. RESULTS & DISCUSSION

### A. MCQ

- <u>QUESTION GENERATION RESULTS</u>

Table 3 shows the MCQ generation performance of the evaluated models across three grade ranges and the All Grades setting, comparing raw and normalized inputs. Overall, the results indicate that the effect of normalization is highly model-dependent and grade-dependent, rather than consistently beneficial across all evaluation settings. While some models exhibit relatively stable performance following normalization, others show substantial changes in lexical similarity, suggesting differences in their sensitivity to input representation.

For Gemma-3-4b and MedGemma-4b, normalization produces relatively moderate and inconsistent changes across the grade ranges. For Gemma-3-4b, normalization improves ROUGE-L from 0.4206 to 0.4396 for Grades 1–2 and from 0.4486 to 0.4568 for Grades 5–6, while slightly reducing it from 0.4442 to 0.4349 for Grades 3–4. A similar pattern is observed for MedGemma-4b, where normalization improves ROUGE-L for Grades 1–2 (0.4319 → 0.4349) and Grades 5–6 (0.4311 → 0.4573), but reduces it for Grades 3–4 (0.4430 → 0.4275). BLEU follows a comparable pattern for MedGemma-4b, with the largest improvement occurring for Grades 5–6 (0.4350 → 0.4920). These findings suggest that normalization has a relatively limited but grade-dependent effect on the two models. Their comparatively stable ROUGE scores also indicate greater consistency across preprocessing conditions than that observed for the smaller models.

A particularly notable pattern is observed for Gemma-2-2b. Unlike the other models, its performance varies considerably across grade ranges and preprocessing conditions. With raw inputs, the model obtains relatively high ROUGE-L scores for Grades 1–2 and Grades 3–4 (0.6102 and 0.8564, respectively), but its performance drops substantially for Grades 5–6 (0.0276) and All Grades (0.0028). Following normalization, ROUGE-L increases for Grades 1–2 (0.6102 → 0.7708), while decreasing for Grades 3–4 (0.8564 → 0.7048). For Grades 5–6 and All Grades, normalization produces increases in ROUGE-L from 0.0276 to 0.0327 and from 0.0028 to 0.2838, respectively. However, BLEU does not consistently follow these changes. For example, BLEU decreases for Grades 1–2 (0.3776 → 0.2194), Grades 3–4 (0.2484 → 0.1308), and All Grades (0.1447 → 0.1294) following normalization. This variation suggests that Gemma-2-2b is particularly sensitive to the input representation and that normalization does not necessarily improve the different aspects of lexical similarity captured by ROUGE and BLEU.

Qwen1.5-1.8B exhibits a different and particularly interesting behavior. Under raw input conditions, its ROUGE-L scores remain close to zero across all grade ranges, indicating very limited lexical overlap with the reference outputs. However, normalization produces substantial increases for the individual grade groups, with ROUGE-L reaching 0.9822, 0.9671, and 0.8725 for Grades 1–2, Grades 3–4, and Grades 5–6, respectively. Despite these large increases, BLEU remains considerably lower, reaching 0.3687, 0.2655, and 0.2213 for the corresponding grade ranges. This divergence is particularly evident for Grades 1–2, where normalized ROUGE-L reaches 0.9822 while BLEU remains at 0.3687. A different pattern emerges when considering All Grades, where normalization produces only a small improvement in ROUGE-L (0.0096 → 0.0139) and BLEU (0.1367 → 0.1559). The substantial discrepancy between the individual grade-level results and the All-Grades results suggests that the model's performance may be sensitive to the composition and distribution of the evaluation data. These results warrant further investigation into the interaction between normalization, input characteristics, and reference alignment.

The results also reveal differences in model performance across grade ranges. For normalized Gemma-3-4b, ROUGE-L and BLEU increase from Grades 1–2 to Grades 5–6 (0.4396 → 0.4568 and 0.4506 → 0.4796, respectively). MedGemma-4b exhibits a similar pattern, with its strongest normalized performance occurring for Grades 5–6 (ROUGE-L:

0.4573; BLEU: 0.4920). In contrast, Gemma-2-2b shows substantial variation across grade ranges, while Qwen1.5-1.8B achieves its highest normalized ROUGE-L for Grades 1–2. These differences may reflect variations in linguistic structure, vocabulary, content distribution, or reference question characteristics across grade levels. However, the results do not establish that content from one grade range is inherently easier to process or generate questions for, as differences in the underlying datasets may also contribute to the observed patterns.

When considering all grades together, normalization generally reduces the ROUGE-L and BLEU scores of Gemma-3-4b and MedGemma-4b. For Gemma-3-4b, ROUGE-L decreases from 0.4647 to 0.4242, while BLEU decreases from 0.4759 to 0.4385. Similarly, MedGemma-4b shows a decrease in ROUGE-L from 0.4473 to 0.4256 and in BLEU from 0.4815 to 0.4471. In contrast, Qwen1.5-1.8B exhibits a small improvement in both metrics, although its absolute ROUGE-L remains substantially lower than that of the other models. Gemma-2-2b demonstrates an increase in ROUGE-L from 0.0028 to 0.2838, accompanied by a decrease in BLEU from 0.1447 to 0.1294. These aggregate results further demonstrate that the effect of normalization observed within individual grade groups does not necessarily carry over to the combined evaluation setting.

Taken together, these findings suggest that normalization should not be regarded as a universally beneficial preprocessing step for MCQ generation. Instead, its impact depends on the interaction between the model, grade range, and characteristics of the input material. Gemma-3-4b and MedGemma-4b demonstrate comparatively stable performance across evaluation settings, whereas Gemma-2-2b and Qwen1.5-1.8B exhibit substantially greater variation, particularly in their ROUGE scores. The differences between ROUGE-L and BLEU also highlight an important limitation of relying exclusively on lexical-overlap metrics. In the context of educational MCQ generation, high lexical similarity does not necessarily indicate that generated questions are factually correct, pedagogically appropriate, aligned with the intended grade level, or accompanied by effective distractors. Therefore, these automatic metrics should be complemented by additional evaluation of question correctness, answer validity, distractor quality, and pedagogical suitability to provide a more comprehensive assessment of MCQ generation performance.

| **Grade Range** | **Text** | **Model** | **ROUGE-1** | **ROUGE-2** | **ROUGE-L** | **BLEU** |
|---|---|---|---|---|---|---|
| **Grades 1–2** | Raw | Gemma-3-4b | 0.4864 | 0.1863 | 0.4206 | 0.4345 |
| | | MedGemma-4b | 0.5100 | 0.2122 | 0.4319 | 0.4586 |
| | | Qwen1.5-1.8B | 0.00000 | 0.00000 | 0.00000 | 0.162824 |
| | | Gemma-2-2b | 0.6102 | 0.3341 | 0.6102 | 0.3776 |
| | Normalized | Gemma-3-4b | 0.4959 | 0.1992 | 0.4396 | 0.4506 |
| | | MedGemma-4b | 0.4964 | 0.2032 | 0.4349 | 0.4561 |
| | | Qwen1.5-1.8B | 0.982156 | 0.961736 | 0.982156 | 0.368673 |
| | | Gemma-2-2b | 0.7708 | 0.6980 | 0.7708 | 0.2194 |
| **Grades 3–4** | Raw | Gemma-3-4b | 0.5303 | 0.2202 | 0.4442 | 0.4772 |
| | | MedGemma-4b | 0.5155 | 0.2047 | 0.4430 | 0.4699 |
| | | Qwen1.5-1.8B | 0.005769 | 0.00206 | 0.005769 | 0.147609 |
| | | Gemma-2-2b | 0.8652 | 0.6907 | 0.8564 | 0.2484 |
| | Normalized | Gemma-3-4b | 0.4997 | 0.2057 | 0.4349 | 0.4496 |
| | | MedGemma-4b | 0.4889 | 0.2004 | 0.4275 | 0.4465 |
| | | Qwen1.5-1.8B | 0.968639 | 0.933870 | 0.967095 | 0.265545 |
| | | Gemma-2-2b | 0.7032 | 0.5967 | 0.7048 | 0.1308 |
| **Grades 5–6** | Raw | Gemma-3-4b | 0.5141 | 0.2162 | 0.4486 | 0.4694 |
| | | MedGemma-4b | 0.4837 | 0.1823 | 0.4311 | 0.4350 |
| | | Qwen1.5-1.8B | 0.009644 | 0.002110 | 0.009644 | 0.136682 |
| | | Gemma-2-2b | 0.0276 | 030259 | 0.0276 | 0.0340 |
| | Normalized | Gemma-3-4b | 0.5205 | 0.2351 | 0.4568 | 0.4796 |
| | | MedGemma-4b | 0.5275 | 0.2253 | 0.4573 | 0.4920 |
| | | Qwen1.5-1.8B | 0.890868 | 0.743215 | 0.872499 | 0.221306 |
| | | Gemma-2-2b | 0.0421 | 0.0040 | 0.0327 | 0.0463 |
| **All Grades** | Raw | Gemma-3-4b | 0.5282 | 0.2325 | 0.4647 | 0.4759 |
| | | MedGemma-4b | 0.5317 | 0.2069 | 0.4473 | 0.4815 |
| | | Qwen1.5-1.8B | 0.009644 | 0.002110 | 0.009644 | 0.136682 |
| | | Gemma-2-2b | 0.0028 | 0.0000 | 0.0028 | 0.1447 |
| | Normalized | Gemma-3-4b | 0.4845 | 0.1955 | 0.4242 | 0.4385 |
| | | MedGemma-4b | 0.4953 | 0.1930 | 0.4256 | 0.4471 |
| | | Qwen1.5-1.8B | 0.013800 | 0.005226 | 0.013890 | 0.155944 |
| | | Gemma-2-2b | 0.2838 | 0.1656 | 0.2838 | 0.1294 |

*Table 3: MCQ Generation Results*

- ANSWER GENERATION RESULTS

Table 4 presents the answer generation accuracy of the evaluated language models across different grade ranges. In contrast to the lexical-overlap results reported in Table 3, the accuracy results show consistently high performance for the Gemma-based models, while Qwen1.5-1.8B exhibits comparatively lower accuracy across all grade levels. Overall, the results indicate that the evaluated models are generally capable of generating correct answers, although performance varies according to both model and grade range.

Gemma-3-4b achieves consistently high answer accuracy across all grade ranges. Its accuracy is 96.2% for Grades 1–2, increases to 97.7% for Grades 3–4, and remains 96.2% for Grades 5–6. When all grades are considered jointly, the model reaches 100% accuracy, representing the highest aggregate accuracy among the evaluated models. The relatively small variation across grade ranges suggests that answer generation remains stable across different levels of educational content.

Gemma-2-2b also demonstrates strong answer generation performance, achieving 100% accuracy for Grades 1–2 and 97.7% for Grades 3–4. Its accuracy decreases slightly to 96.2% for Grades 5–6 and 97% when all grades are combined. Thus, although Gemma-2-2b has slightly lower aggregate accuracy than Gemma-3-4b, the difference is small, and both models maintain accuracy above 96% across individual grade ranges. These results suggest that answer generation can remain highly reliable even for the smaller model, despite the substantial differences observed between models in the MCQ generation metrics in Table 3.

In contrast, Qwen1.5-1.8B shows consistently lower answer accuracy. Its performance decreases from 90.38% for Grades 1–2 to 92.05% for Grades 3–4 and 84.81% for Grades 5–6, with an overall accuracy of 93.00%. The lowest result is observed for Grades 5–6, where the gap relative to the Gemma models is particularly pronounced. This pattern suggests that the model's answer generation may be more sensitive to the complexity or characteristics of the educational content at higher grade levels. However, the table alone does not establish the specific cause of this difference.

The grade-level variation is also noteworthy. For Gemma-3-4b, accuracy remains within a narrow range of 96.2–97.7% across the three grade ranges. Gemma-2-2b follows a similar pattern, with only a modest decrease from 100% for Grades 1–2 to 96.2% for Grades 5–6. Qwen1.5-1.8B, however, exhibits a larger variation, with its accuracy dropping to 84.81% for Grades 5–6. This contrast indicates that answer-generation robustness across grade levels differs substantially among models.

Taken together, Table 4 provides evidence that answer generation accuracy is generally high for the Gemma models and comparatively lower for Qwen1.5-1.8B. Importantly, these results complement the findings in Table 3: a model's lexical similarity to reference MCQs does not necessarily correspond directly to its ability to generate correct answers. For example, Qwen1.5-1.8B shows substantially lower ROUGE-L scores in Table 3 while maintaining relatively high answer accuracy in Table 4. This distinction highlights the importance of evaluating MCQ generation using multiple dimensions rather than relying solely on text-overlap metrics.

Finally, although the accuracy values are high, answer accuracy alone does not capture all aspects of MCQ quality. A generated question may have the correct answer while still containing ambiguous wording, implausible distractors, inappropriate difficulty, or poor alignment with the source material.

| Grade Range | Gemma-3-4b | Gemma-2-2b | Qwen1.5-1.8B |
|---|---|---|---|
| **Grades 1–2** | 96.2% | 100% | 90.38% |
| **Grades 3–4** | 97.7% | 97.7% | 92.05% |
| **Grades 5–6** | 96.2% | 96.2% | 84.81% |
| **All Grades** | 100% | 97% | 93.00% |

*Table 4: Answer Generation Accuracy Results*

## B. SCIENTITIFC EXPERIMENTS GENERATION RESULTS

Table 5 presents the results of scientific experiment generation across different grade ranges, comparing model performance on raw and normalized educational text. Unlike tasks that focus on generating short answers or multiple-choice questions, scientific experiment generation requires models to produce descriptions that preserve the lexical content and structure of reference experiments. The results reveal considerable variation across models and grade levels, indicating that the effect of text normalization depends strongly on the underlying model and the educational content being processed.

As for the effect of normalization, normalization has a particularly positive effect on Gemma-3-4b. Across all grade ranges, the model maintains strong ROUGE performance while demonstrating substantial improvements in BLEU. For Grades 1–2, ROUGE increases from 0.9654 to 0.9748, while BLEU rises markedly from 0.3416 to 0.8926. For Grades 3–4, ROUGE increases from 0.9665 to 0.9756 and BLEU from 0.3328 to 0.5377. The strongest normalized ROUGE result is obtained for Grades 5–6, reaching 0.9883 compared with 0.9315 under raw input, while BLEU increases from 0.3313 to 0.8542. At the aggregate level, normalized input also improves Gemma-3-4b, with ROUGE increasing from 0.9431 to 0.9508 and BLEU from 0.3563 to 0.5319. These results suggest that normalization substantially improves the lexical alignment of generated scientific experiments with the reference descriptions for Gemma-3-4b, particularly in terms of BLEU.

MEDGemma-3-4b also achieves strong performance, although the effect of normalization is less consistent across individual grade ranges. For Grades 1–2, normalization produces a small decrease in ROUGE (1.0000 → 0.9937), while BLEU increases from 0.4892 to 0.6243. For Grades 3–4, ROUGE increases from 0.9524 to 0.9781, whereas BLEU decreases slightly from 0.3578 to 0.3406. For Grades 5–6, both metrics decrease after normalization, with ROUGE changing from 0.9679 to 0.9557 and BLEU from 0.5016 to 0.3236. Nevertheless, when all grades are considered together, the normalized setting achieves 0.9701 ROUGE and 0.5384 BLEU, compared with 0.7801 and 0.4827 for raw input. This indicates a substantial improvement in the aggregate results, despite the heterogeneous effects observed across individual grade ranges.

Qwen1.5-1.8B demonstrates a mixed response to normalization. For Grades 1–2, its ROUGE and BLEU scores remain nearly unchanged, with ROUGE decreasing slightly from 0.9837 to 0.9822 and BLEU from 0.3700 to 0.3687. For Grades 3–4, normalization improves ROUGE from 0.9540 to 0.9671, although BLEU decreases from 0.2854 to 0.2655. A different pattern is observed for Grades 5–6, where both ROUGE and BLEU decrease, from 0.9157 to 0.8725 and from 0.2517 to 0.2213, respectively. At the aggregate level, normalization also results in a small decrease in ROUGE (0.9607 → 0.9591) and BLEU (0.2946 → 0.2802). These results suggest that Qwen1.5-1.8B is relatively insensitive to normalization for lower-grade content but experiences performance degradation for some higher-grade configurations.

The results for StableLM-2-1.6B show relatively stable performance between raw and normalized input. For Grades 1–2, ROUGE decreases slightly from 0.9899 to 0.9844, while BLEU increases from 0.3608 to 0.3651. When all grades are considered, ROUGE increases marginally from 0.9332 to 0.9356, while BLEU remains almost unchanged (0.3461 → 0.3462). These results indicate that normalization has a limited effect on StableLM-2-1.6B, suggesting that its generation performance is relatively consistent across the two preprocessing conditions.

The results for BLOOM further illustrate differences in performance across model configurations. Under raw input for all grades, BLOOM-560M achieves a ROUGE score of 0.9363 but a considerably lower BLEU score of 0.0641. In contrast, BLOOM-1.7B obtains a ROUGE score of 0.8597 and a BLEU score of 0.7006. This substantial difference between the two metrics suggests that the models exhibit different patterns of lexical overlap with the reference experiment descriptions. Although BLOOM-560M demonstrates relatively high ROUGE, its low BLEU indicates weaker alignment in terms of exact n-gram sequences. BLOOM-1.7B, on the other hand, achieves substantially higher BLEU despite its lower ROUGE score. These findings highlight the importance of considering multiple evaluation metrics when comparing generated scientific experiments.

TinyLlama-1.1B demonstrates relatively strong ROUGE performance under raw input but considerably lower BLEU scores. For Grades 1–2, it achieves 0.9410 ROUGE and 0.0725 BLEU, while the corresponding normalized results are 0.9357 and 0.0677. When all grades are considered, ROUGE decreases from 0.9410 to 0.8449, accompanied by a BLEU decrease from 0.0726 to 0.0658. This suggests that normalization does not improve the lexical alignment of TinyLlama-generated experiments and may negatively affect its performance in the aggregate setting.

For Falcon-H1-0.5B, the available results for Grades 5–6 show virtually identical performance under raw and normalized input. ROUGE remains at approximately 0.7872, while BLEU changes only marginally from 0.1375 to 0.1375. This indicates that normalization has almost no measurable effect on the model's performance for this grade range. However, its relatively low ROUGE and BLEU scores compared with Gemma-3-4b and MEDGemma-3-4b suggest weaker lexical alignment with the reference experiments.

As for grade-level effects, the results indicate that scientific experiment generation performance varies across educational levels and models. Gemma-3-4b maintains consistently strong ROUGE performance across all grade ranges and achieves its highest normalized ROUGE score for Grades 5–6 (0.9883). Its normalized BLEU also remains high across grade levels, with scores of 0.8926, 0.5377, and 0.8542 for Grades 1–2, 3–4, and 5–6, respectively. MEDGemma-3-4b demonstrates its strongest normalized BLEU for Grades 1–2 (0.6243), followed by lower scores for Grades 3–4 and Grades 5–6. Qwen1.5-1.8B also exhibits a decrease in ROUGE and BLEU for Grades 5–6 under normalized input. These variations may reflect differences in vocabulary, procedural complexity, and the structure of scientific concepts presented at different educational levels. However, the results alone do not establish whether these differences are attributable to grade-level complexity or other characteristics of the underlying datasets.

Overall, the findings indicate that normalization has a strongly model-dependent effect on scientific experiment generation. Gemma-3-4b exhibits the most consistent improvement, particularly in BLEU, while MEDGemma-3-4b achieves strong aggregate normalized performance despite mixed results across individual grade ranges. Qwen1.5-1.8B demonstrates relatively small but predominantly negative changes, whereas StableLM-2-1.6B and Falcon-H1-0.5B show comparatively stable performance across preprocessing conditions. TinyLlama-1.1B does not benefit from normalization, while the BLOOM results highlight substantial differences between ROUGE and BLEU. Taken together, these findings suggest that text normalization should be evaluated separately for each model and educational setting rather than assumed to provide a universal improvement in scientific experiment generation.

| Grade Range | Text | Model | ROUGE 1 | ROUGE 2 | ROUGE | BLEU |
|---|---|---|---|---|---|---|
| Grades 1–2 | Raw | Gemma-3-4b | 0.9665 | 0.9600 | 0.9654 | 0.3416 |
| | | MEDGemma-3-4b | 1.0000 | 1.0000 | 1.0000 | 0.4892 |
| | | Qwen1.5-1.8B | 0.983720 | 0.963049 | 0.983720 | 0.369998 |
| | | BLOOM-560M | 0.966578 | 0.913636 | 0.958447 | 0.059711 |
| | | StableLM-2-1.6B | 0.989895 | 0.979390 | 0.989895 | 0.360847 |
| | | TinyLlama-1.1B | 0.941026 | 0.862255 | 0.941026 | 0.072517 |
| | **Normalized** | Gemma-3-4b | 0.9907 | 0.9863 | 0.9748 | 0.8926 |
| | | MEDGemma-3-4b | 0.9946 | 0.9881 | 0.9937 | 0.6243 |
| | | Qwen1.5-1.8B | 0.982156 | 0.961736 | 0.982156 | 0.368673 |
| | | TinyLlama-1.1B | 0.935654 | 0.855576 | 0.935654 | 0.067717 |
| | | StableLM-2-1.6B | 0.987711 | 0.969241 | 0.984443 | 0.365118 |
| **Grades 3–4** | **Raw** | Gemma-3-4b | 0.9681 | 0.9380 | 0.9665 | 0.3328 |
| | | MEDGemma-3-4b | 0.9544 | 0.9125 | 0.9524 | 0.3578 |
| | | Qwen1.5-1.8B | 0.955434 | 0.917106 | 0.954005 | 0.285354 |
| | **Normalized** | Gemma-3-4b | 0.9770 | 0.9523 | 0.9756 | 0.5377 |
| | | MEDGemma-3-4b | 0.9689 | 0.9337 | 0.9781 | 0.3406 |
| | | Qwen1.5-1.8B | 0.968639 | 0.933870 | 0.967095 | 0.265545 |
| **Grades 5–6** | **Raw** | Gemma-3-4b | 0.9407 | 0.8827 | 0.9315 | 0.3313 |
| | | MEDGemma-3-4b | 0.9693 | 0.9397 | 0.9679 | 0.5016 |
| | | Qwen1.5-1.8B | 0.929756 | 0.854077 | 0.915674 | 0.251683 |
| | | Falcon-H1-0.5B | 0.791532 | 0.717278 | 0.78721 | 0.13752 |
| | **Normalized** | Gemma-3-4b | 0.9429 | 0.8877 | 0.9883 | 0.8542 |
| | | MEDGemma-3-4b | 0.9631 | 0.9259 | 0.9557 | 0.3236 |
| | | Qwen1.5-1.8B | 0.890868 | 0.743215 | 0.872499 | 0.221306 |
| | | Falcon-H1-0.5B | 0.791532 | 0.717278 | 0.78721 | .137547 |
| **All Grades** | **Raw** | Gemma-3-4b | 0.9440 | 0.9167 | 0.9431 | 0.3563 |
| | | MEDGemma-3-4b | 0.9551 | 0.9192 | 0.7801 | 0.4827 |
| | | Qwen1.5-1.8B | 0.963263 | 0.925652 | 0.960683 | 0.294598 |
| | | BLOOM-560M | 0.944868 | 0.873796 | 0.936292 | 0.064059 |
| | | BLOOM-1.7B | 0.8610 | 0.8427 | 0.8597 | 0.7006 |
| | | StableLM-2-1.6B | 0.935062 | 0.869488 | 0.933240 | 0.346096 |
| | | TinyLlama-1.1B | 0.941026 | 0.862255 | 0.941026 | 0.072571 |
| | **Normalized** | Gemma-3-4b | 0.9542 | 0.9277 | 0.9508 | 0.5319 |
| | | MEDGemma-3-4b | 0.9758 | 0.9498 | 0.9701 | 0.5384 |
| | | Qwen1.5-1.8B | 0.959600 | 0.923506 | 0.959112 | 0.280195 |
| | | TinyLlama-1.1B | 0.847391 | 0.677191 | 0.844891 | 0.065845 |
| | | StableLM-2-1.6B | 0.937392 | 0.874873 | 0.935571 | 0.346191 |

*Table 5: Scientific Experiments Results*

# VII. HUMAN EVALUATION

As for scientific experiment generation or MCQ generation, an experiment or question may use different wording from the reference while still describing the same scientific procedure, variables, materials, questions, or expected outcome. Consequently, high lexical overlap is not necessarily equivalent to scientific correctness or experimental usefulness. Conversely, a low lexical-overlap score does not necessarily indicate that the generated experiment is scientifically invalid. Therefore, we have conducted a human evaluation as illustrated in this section.

## A. HUMAN EVALUATION PROTOCOL

Five-criterion human evaluation was conducted for 50 random selected samples for each tuning experiment.

Scientific Experiments Generation criteria are:

**Factual Accuracy:** is the experiment scientifically sound, allowing pedagogical simplification?

- **Structural Integrity:** Does the output contain the required five sections: Goal, Tools, Steps, Observations, and Conclusion?
- **Logical Coherence:** Are the sections logically connected to each other?
- **Practical Value:** Can the experiment actually be performed at the target grade level?
- **Language Fluency:** Does the output read as natural Arabic, free of English words, code, HTML, or stray symbols?
- **Overall:** The Overall column reports the average of the five raw scores.

MCQ Generation criteria are:

- **Question:** The output starts with "السؤال" word with the question content.
- **4 Options:** There are four distinct answers options labelled as أ/ب/ج/د
- **No Duplicated:** Options are not lexically identical
- **Only Arabic:** Generated question and answer option are all pure Arabic with no English words or code
- **Content Quality:** The question is relevant, the correct answer is among the options, distractors are plausible, and no hallucinated facts.
- **Score:** The Score column is the average of the five raw fractions.

## B. HUMAN EVALUATION RESULTS

Tables 6 and 7 provide a complementary human evaluation of the generated **MCQs and scientific experiments**, respectively. While the automatic evaluation measures lexical similarity and answer accuracy, the human evaluation assesses whether the generated outputs satisfy task-specific quality requirements. Overall, the human judgments reveal substantial differences between models and, importantly, show that improvements in automatic metrics do not necessarily translate into improvements in human-perceived quality.

-MCQ GENERATION

Table 6 evaluates generated MCQs according to five criteria: question validity, completeness of the four options, absence of duplicated options, Arabic-only generation, and content quality. The results show that Gemma-3-4b and MedGemma-4b maintain consistently high human-evaluated quality across grade ranges**.** Under raw input, Gemma-3-4b achieves scores of 99.2%, 99.6%, and 96.4% for Grades 1–2, 3–4, and 5–6, respectively. MedGemma-4b exhibits a similarly high performance, obtaining 99.2%, 97.6%, and 98.8%. These results indicate that both models generally satisfy the structural requirements of MCQ generation while producing content judged to be appropriate by the human evaluators.

For Gemma-3-4b, normalization has only a modest effect on human-evaluated MCQ quality. The overall score changes from 99.2% to 98.4% for Grades 1–2, from 99.6% to 98.8% for Grades 3–4, and from 96.4% to 97.2% for Grades 5–6. Thus, the model remains highly consistent regardless of the input representation. MedGemma-4b shows a similar pattern, with normalized scores of 97.6%, 98.4%, and 98.0% across the three grade ranges.

This stability is notable because it indicates that normalization does not substantially disrupt the ability of these models to produce structurally valid MCQs.

The behavior of Qwen1.5-1.8B is substantially more variable. Under raw input, its scores range from 82.0% to 88.7%, whereas normalized input produces scores of 82.8%, 42.0%, and 84.8% for Grades 1–2, 3–4, and 5–6, respectively. The particularly low score for Grades 3–4 is associated with severe structural and language-related failures: the normalized model receives 0/50 for both the question and four-option criteria and only 5/50 for the Arabic-only criterion, despite receiving 50/50 for the absence of duplicated options. This suggests that the degradation is not simply a reduction in general content quality but is concentrated in the model's ability to produce valid, appropriately formatted Arabic MCQs under this particular setting.

Gemma-2-2b presents an interesting contrast. Its raw performance is lower for the individual grade ranges, with scores of 74%, 72%, and 76%, but its normalized scores increase to 90%, 93%, and 97%. The aggregate normalized score is also high at 97%. This substantial improvement indicates that normalization can be highly beneficial for MCQ generation with this model from a human-evaluation perspective. It also demonstrates that the effect of normalization cannot be generalized across models: while it has little effect on Gemma-3-4b and MedGemma-4b, it substantially improves Gemma-2-2b while producing a severe degradation for Qwen1.5-1.8B in one grade range.

An important observation is the difference between automatic and human evaluation. The automatic results in the previous table showed substantial variation in ROUGE and BLEU across models and preprocessing conditions, whereas the human evaluation shows that some models with relatively different lexical-overlap behavior can nevertheless produce highly acceptable MCQs. This supports the argument that lexical similarity alone is insufficient for evaluating educational question generation. Human evaluation captures structural properties and content characteristics that are not necessarily reflected in n-gram-based metrics.

| **Grade Range** | **Text** | **Model** | **"Question:"** | **4 Options** | **No Duplicated** | **Only Arabic** | **Content Quality** | **Score** |
|---|---|---|---|---|---|---|---|---|
| **Grades 1–2** | **Raw** | Gemma-3-4b | 50/50 | 50/50 | 50/50 | 50/50 | 48/50 | **99.2%** |
| | | MedGemma-4b | 50/50 | 50/50 | 50/50 | 50/50 | 48/50 | **99.2%** |
| | | Qwen1.5-1.8B | 50/50 | 50/50 | 49/50 | 36/50 | 20/50 | **82.0%** |
| | | Gemma-2-2b | 50/50 | 39/50 | 31/50 | 34/50 | 33/50 | **74%** |
| | **Normalized** | Gemma-3-4b | 50/50 | 50/50 | 50/50 | 50/50 | 46/50 | **98.4%** |
| | | MedGemma-4b | 50/50 | 50/50 | 50/50 | 50/50 | 44/50 | **97.6%** |
| | | Qwen1.5-1.8B | 50/50 | 50/50 | 50/50 | 42/50 | 15/50 | **82.8%** |
| | | Gemma-2-2b | 30/30 | 30/30 | 29/30 | 24/30 | 25/30 | **90%** |
| **Grades 3–4** | **Raw** | Gemma-3-4b | 50/50 | 50/50 | 50/50 | 50/50 | 49/50 | **99.6%** |
| | | MedGemma-4b | 50/50 | 50/50 | 50/50 | 50/50 | 44/50 | **97.6%** |
| | | Qwen1.5-1.8B | 50/50 | 50/50 | 49/50 | 45/50 | 17/50 | **84.4%** |
| | | Gemma-2-2b | 50/50 | 38/50 | 29/50 | 32/50 | 34/50 | **72%** |
| | **Normalized** | Gemma-3-4b | 50/50 | 50/50 | 50/50 | 50/50 | 47/50 | **98.8%** |
| | | MedGemma-4b | 50/50 | 50/50 | 50/50 | 50/50 | 46/50 | **98.4%** |
| | | Qwen1.5-1.8B | 0/50 | 0/50 | 50/50 | 5/50 | 0/50 | **42.0%** |
| | | Gemma-2-2b | 30/30 | 30/30 | 30/30 | 25/30 | 24/30 | **93%** |
| **Grades 5–6** | **Raw** | Gemma-3-4b | 50/50 | 50/50 | 50/50 | 47/50 | 44/50 | **96.4%** |
| | | MedGemma-4b | 50/50 | 50/50 | 50/50 | 49/50 | 48/50 | **98.8%** |
| | | Qwen1.5-1.8B | 50/50 | 50/50 | 49/50 | 46/50 | 23/50 | **87.2%** |
| | | Gemma-2-2b | 50/50 | 41/50 | 34/50 | 31/50 | 35/50 | **76%** |
| | **Normalized** | Gemma-3-4b | 50/50 | 50/50 | 50/50 | 48/50 | 45/50 | **97.2%** |
| | | MedGemma-4b | 50/50 | 50/50 | 50/50 | 47/50 | 48/50 | **98.0%** |
| | | Qwen1.5-1.8B | 50/50 | 50/50 | 50/50 | 43/50 | 19/50 | **84.8%** |
| | | Gemma-2-2b | 30/30 | 30/30 | 30/30 | 28/30 | 27/30 | **97%** |
| **All Grades** | **Raw** | Gemma-3-4b | 50/50 | 50/50 | 50/50 | 48/50 | 45/50 | **94.7%** |
| | | MedGemma-4b | 50/50 | 50/50 | 50/50 | 49/50 | 48/50 | **98.8%** |
| | | Qwen1.5-1.8B | 30/30 | 30/30 | 29/30 | 30/30 | 14/30 | **88.7%** |
| | | Gemma-2-2b | 30/30 | 30/30 | 30/30 | 30/30 | 29/30 | **100%** |
| | **Normalized** | Gemma-3-4b | 30/30 | 30/30 | 30/30 | 30/30 | 28/30 | **98.7%** |
| | | MedGemma-4b | 50/50 | 50/50 | 50/50 | 49/50 | 48/50 | **98.8%** |
| | | Qwen1.5-1.8B | 30/30 | 30/30 | 30/30 | 22/30 | 4/30 | **77.3%** |
| | | Gemma-2-2b | 30/30 | 30/30 | 30/30 | 28/30 | 28/30 | **97%** |

*Table 6: MCQ Generation Human Evaluation Results*

## -SCIENTITIFC EXPERIMENTS GENERATION RESULTS

Table 7 evaluates the generated scientific experiments according **to** factual accuracy, structural integrity, logical coherence, practical value, and language fluency. Compared with MCQ generation, the human evaluation reveals substantially greater variation across models and preprocessing conditions. This is expected because scientific experiment generation requires the model to simultaneously satisfy scientific, procedural, logical, practical, and linguistic constraints.

Under raw input, Gemma-3-4b achieves the highest aggregate human-evaluation score among the evaluated models, with an overall score of 80.8%. Its performance is relatively strong across all criteria, particularly structural integrity (100%) and language fluency (100%). However, its logical coherence score is lower at 60%, indicating that generating a structurally complete experiment does not necessarily guarantee that the experimental steps form a fully coherent procedure. Factual accuracy is also 70%, suggesting room for improvement in scientific correctness.

MedGemma-4b obtains an aggregate score of 75.2% under raw input. Its strongest dimensions are structural integrity (100%) and language fluency (96%), while factual accuracy, logical coherence, and practical value are comparatively lower. This pattern suggests that the model generally produces well-formed and linguistically fluent experiments but may require additional evaluation or constraints to ensure scientific and procedural quality.

Qwen1.5-1.8B achieves an aggregate score of 65% under raw input. Its performance is relatively strong in structural integrity (90%) and language fluency (74%), but factual accuracy (52%) and practical value (44%) are lower. These results indicate that the model can produce reasonably structured experimental descriptions while having greater difficulty ensuring that the proposed experiments are scientifically accurate and practically meaningful.

The largest difference under raw input is observed for Gemma-2-2b, which obtains an aggregate score of only 22%. Its scores are particularly low for factual accuracy (16%), logical coherence (20%), and practical value (14%). This contrasts with its relatively strong MCQ performance after normalization in Table 6 and demonstrates that performance is strongly task-dependent. A model that can successfully generate structured MCQs may still struggle with the more demanding scientific reasoning and procedural requirements involved in experiment generation.

As for the effect of normalization on experiment quality, the human evaluation shows that normalization has markedly different effects on scientific experiment generation compared with MCQ generation. For Gemma-3-4b, the overall score decreases from 80.8% to 76.0% after normalization. Although structural integrity remains at 100% and language fluency remains high at 100%, factual accuracy decreases from 70% to 66%, logical coherence from 60% to 54%, and practical value from 74% to 60%. Thus, normalization does not improve the overall human-perceived quality of experiments for this model and may negatively affect several substantive dimensions. MedGemma-4b exhibits a similar pattern, with its overall score decreasing from 75.2% to 72.0%. The largest changes occur in factual accuracy (60% → 56%), practical value (64% → 52%), and language fluency (96% → 100%, which improves). This suggests that normalization does not uniformly affect all aspects of experiment generation: linguistic fluency can remain strong or improve while scientific and practical dimensions decline. The most pronounced change is observed for Gemma-2-2b. Its overall score increases dramatically from 22% under raw input to 66% under normalized input. Improvements occur across all dimensions, including factual accuracy (16% → 56%), structural integrity (36% → 80%), logical coherence (20% → 62%), practical value (14% → 38%), and language fluency (40% → 70%). This provides strong human-evaluation evidence that normalization can substantially improve the quality of scientific experiment generation for this model. In contrast, Qwen1.5-1.8B shows a substantial decline after normalization, with its overall score falling from 65% to 25%. The decrease is particularly pronounced in factual accuracy (52% → 10%), logical coherence (64% → 10%), and language fluency (74% → 10%). Although practical value increases substantially from 44% to 96%, this improvement is accompanied by severe degradation in most of the other dimensions. This result highlights that normalization can introduce undesirable changes in generation behavior even when it improves an individual evaluation dimension.

| Grade Range | Text | Model | Factual Accuracy | Structural Integrity | Logical Coherence | Practical Value | Language Fluency | Overall |
|---|---|---|---|---|---|---|---|---|
| Grades 1–2 | Raw | Gemma-3-4b | 70% | 100% | 64% | 74% | 100% | **81.6%** |
| | | MEDGemma-3-4b | 56% | 100% | 48% | 54% | 94% | **70.4%** |
| | | Qwen1.5-1.8B | 36% | 100% | 40% | 82% | 0% | **41%** |
| | | Gemma-2-2b | 14% | 84% | 20% | 12% | 26% | **31.2%** |
| | Normalized | Gemma-3-4b | 54% | 98% | 40% | 44% | 92% | **65.6%** |
| | | MEDGemma-3-4b | 50% | 100% | 44% | 46% | 100% | **68.0%** |
| | | Qwen1.5-1.8B | 70% | 98% | 72% | 90% | 0% | **63%** |
| | | Gemma-2-2b | 38% | 90% | 48% | 52% | 54% | **54%** |
| Grades 3–4 | Raw | Gemma-3-4b | 46% | 92% | 28% | 44% | 94% | **60.8%** |
| | | MEDGemma-3-4b | 36% | 94% | 32% | 34% | 30% | **45.2%** |
| | | Qwen1.5-1.8B | 26% | 86% | 24% | 56% | 0% | **32%** |
| | | Gemma-2-2b | 12% | 42% | 10% | 8% | 48% | **20%** |
| | Normalized | Gemma-3-4b | 74% | 94% | 52% | 80% | 96% | **79.2%** |
| | | MEDGemma-3-4b | 44% | 88% | 38% | 42% | 48% | **52.0%** |
| | | Qwen1.5-1.8B | 28% | 70% | 32% | 8% | 0% | **22%** |
| | | Gemma-2-2b | 22% | 66% | 28% | 20% | 48% | **37%** |
| Grades 5–6 | Raw | Gemma-3-4b | 50% | 100% | 48% | 56% | 100% | **70.8%** |
| | | MEDGemma-3-4b | 30% | 90% | 34% | 32% | 84% | **54.0%** |
| | | Qwen1.5-1.8B | 40% | 96% | 28% | 6% | 8% | **35%** |
| | | Gemma-2-2b | 14% | 52% | 16% | 36% | 36% | **26%** |
| | **Normalized** | Gemma-3-4b | 72% | 100% | 46% | 50% | 100% | **73.6%** |
| | | MEDGemma-3-4b | 24% | 82% | 44% | 40% | 60% | **50.0%** |
| | | Qwen1.5-1.8B | 2% | 28% | 4% | 0% | 0% | **7%** |
| | | Gemma-2-2b | 22% | 38% | 28% | 20% | 36% | **34%** |
| **All Grades** | **Raw** | Gemma-3-4b | 70% | 100% | 60% | 74% | 100% | **80.8%** |
| | | MEDGemma-3-4b | 60% | 100% | 56% | 64% | 96% | **75.2%** |
| | | Qwen1.5-1.8B | 52% | 90% | 64% | 44% | 74% | **65%** |
| | | Gemma-2-2b | 16% | 36% | 20% | 14% | 40% | **22%** |
| | **Normalized** | Gemma-3-4b | 66% | 100% | 54% | 60% | 100% | **76.0%** |
| | | MEDGemma-3-4b | 56% | 100% | 52% | 52% | 100% | **72.0%** |
| | | Qwen1.5-1.8B | 10% | 78% | 10% | 96% | 10% | **25%** |
| | | Gemma-2-2b | 56% | 80% | 62% | 38% | 70% | **66%** |

*Table 7: Scientific Experiments Human Evaluation Results*

Taken together, Tables 6 and 7 demonstrate that human evaluation provides information that cannot be inferred reliably from automatic lexical metrics alone. For MCQ generation, Gemma-3-4b and MedGemma-4b maintain very high human-evaluated quality across preprocessing conditions, while Gemma-2-2b benefits substantially from normalization. For scientific experiment generation, however, the effect of normalization is more heterogeneous: Gemma-2-2b again benefits strongly, whereas Gemma-3-4b, MedGemma-4b, and particularly Qwen1.5-1.8B experience decreases in their aggregate human-evaluation scores.

The difference between the two tasks is also informative. MCQ generation is dominated by structural and content constraints, such as producing a valid question and four distinct options, whereas scientific experiment generation requires more complex reasoning about scientific facts, procedural structure, logical dependencies, and practical feasibility. Consequently, a model can perform strongly on MCQ generation while exhibiting substantially weaker performance on scientific experiment generation.

Overall, the human evaluation suggests that normalization should be treated as a model- and task-specific design choice rather than a universally beneficial preprocessing strategy. More importantly, the results demonstrate the value of evaluating educational generation systems along multiple dimensions. For MCQs, structural validity, language consistency, and content quality are important; for scientific experiments, factual accuracy, logical coherence, practical value, and structural integrity become particularly important. These human-centered criteria provide a more meaningful assessment of educational generation quality than surface-level textual similarity alone.

# VIII. CONCLUSION

ArSciKids provides a structured, curriculum-grounded foundation for Arabic educational NLP at the primary-school level. Its principal contribution is the combination of authentic Arabic source material, grade labels, explicit generation-oriented schemas, and manual quality control across two complementary tasks: scientific-experiment generation and MCQ generation. The 999-record experiment resource offers relatively even grade coverage and cross-curriculum diversity, while the 2,167-record assessment resource offers a more homogeneous question-generation setting but requires grade-aware evaluation because its distribution is more uneven.

The baseline experiments demonstrate that these resources contain sufficient signal for strong adaptation of Arabic-capable 4B models. More importantly, the human-evaluation results show why dataset utility cannot be established by ROUGE and BLEU alone. Structural compliance and Arabic fluency can approach ceiling while logical coherence, factual grounding, and tool–procedure consistency remain imperfect. The targeted study further shows that explicit consistency interventions can improve the weakest dimensions without substantially changing aggregate lexical scores.

ArSciKids is a useful resource for Arabic educational generation, but it does not make educational generation solved. Its strongest future value lies in enabling controlled research on curriculum-aware generation, field-level consistency, grade-sensitive evaluation, and human-centered quality assessment. Expanding curricular coverage, releasing richer dataset statistics, and strengthening the human-evaluation protocol would further increase its value as a reusable benchmark and dataset for Arabic educational AI.